\pdfoutput=1

\documentclass[11pt]{article}

\usepackage{msp}

\usepackage{times}    
\usepackage{latexsym}

\usepackage[T1]{fontenc} 
\usepackage[utf8]{inputenc} 

\usepackage{microtype} 
\usepackage{hyperref}  
\usepackage{graphicx}  
\usepackage{booktabs}  
\usepackage{amsmath}   
\usepackage{amsfonts}
\usepackage{amssymb}
\usepackage{multirow}

\usepackage{tikz}
\usetikzlibrary{positioning, arrows.meta,shapes.geometric}

\DeclareMathOperator*{\argmax}{arg\,max}

\usepackage{subfigure}

\title{Plausible-Parrots @ \textit{MSP2023}: Enhancing Semantic Plausibility Modeling using Entity and Event Knowledge 
}

\author{Chong Shen \and Chenyue Zhou \\         
Institute for Natural Language Processing, University of Stuttgart, Germany \\ \texttt{\{chong.shen,chenyue.zhou\}@ims.uni-stuttgart.de}}
\begin{document}
\maketitle
\begin{abstract}
In this work, we investigate the effectiveness of injecting external knowledge to a large language model (LLM) to identify semantic plausibility of simple events. Specifically, we enhance the LLM with fine-grained entity types, event types and their definitions extracted from an external knowledge base. These knowledge are injected into our system via designed templates. We also augment the data to balance the label distribution and adapt the task setting to real world scenarios in which event mentions are expressed as natural language sentences. The experimental results show the effectiveness of the injected knowledge on modeling semantic plausibility of events. An error analysis further emphasizes the importance of identifying non-trivial entity and event types.\footnote{Code and data are available at \url{https://github.com/st143575/SemPlaus-plausibleparrots}.}

\end{abstract}

\section{Introduction}
\label{sec:introduction}
Discerning expressions about plausible events from implausible ones is a fundamental element for understanding event semantics. Semantic plausibility modeling is the task of identifying events that are likely to happen but not necessarily attested in a given world \citep{gordon2013reporting}. Previous works have shown the potential of incorporating world knowledge in solving the task, such as physical attributes \citep{wang2018modeling}, lexical hierarchy \citep{porada2021modeling} and degrees of abstractness \citep{eichel2023dataset}. 

Events in the real world are typically expressed by natural language sentences, which possess diverse, complex and dynamic forms. The words constituting an event are often ambiguous. For example, the subject, verb and object of the event \textit{Jobs takes an apple} all have multiple meanings and can lead to misinterpretation of the plausibility of the event. Thus, we hypothesize that the types of the trigger (i.e. verb) and the arguments (e.g., subject and object) of an event are crucial to disambiguate the word meaning and thus help the model better understand the semantic plausibility of the event. Furthermore, single events in the form of \textit{(subject,verb,object)}-triples are inconsistent with the natural language sentences that are input to LLMs during pretraining. This mismatch potentially limits the model's performance.

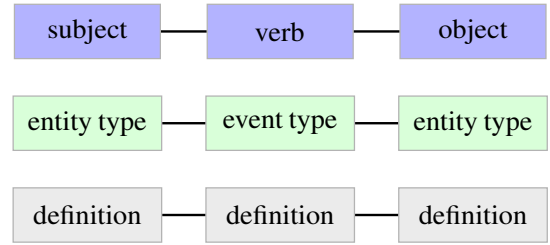
\begin{figure}[tbp]
    \centering
    \scalebox{0.6}{
        \begin{tikzpicture}[scale=1]
    \node[
        shape=rectangle,
        draw=black,
        thick,
        fill=blue!100, 
        opacity=0.3, 
        minimum width=3.2cm, 
        minimum height=1.2cm, 
        text width=2cm, 
        text opacity=1, 
        align=center,
    ] (subject) {\LARGE{subject}};

    \node[
        shape=rectangle,
        draw=black,
        thick,
        fill=blue!100, 
        opacity=0.3, 
        right=1cm of subject,
        minimum width=3.2cm, 
        minimum height=1.2cm, 
        text width=2cm, 
        text opacity=1, 
        align=center,
    ] (verb) {\LARGE{verb}};

    \node[
        shape=rectangle,
        draw=black,
        thick,
        fill=blue!100, 
        opacity=0.3, 
        right=1cm of verb,
        minimum width=3.2cm, 
        minimum height=1.2cm, 
        text width=2cm, 
        text opacity=1, 
        align=center,
    ] (object) {\LARGE{object}};


    \node[
        shape=rectangle,
        draw=black,
        thick,
        fill=green!50, 
        opacity=0.3, 
        below=0.8cm of subject,
        minimum width=3cm, 
        minimum height=1.2cm, 
        text width=3cm, 
        text opacity=1, 
        align=center,
    ] (enttype_subj) {\LARGE{entity type}};

    \node[
        shape=rectangle,
        draw=black,
        thick,
        fill=green!50, 
        opacity=0.3, 
        below=0.8cm of verb,
        minimum width=2cm, 
        minimum height=1.2cm, 
        text width=3cm, 
        text opacity=1, 
        align=center,
    ] (evttype) {\LARGE{event type}};

    \node[
        shape=rectangle,
        draw=black,
        thick,
        fill=green!50, 
        opacity=0.3, 
        below=0.8cm of object,
        minimum width=2cm, 
        minimum height=1.2cm, 
        text width=3cm, 
        text opacity=1, 
        align=center,
    ] (enttype_obj) {\LARGE{entity type}};

    \node[
        shape=rectangle,
        draw=black,
        thick,
        fill=gray!50, 
        opacity=0.3, 
        below=0.8cm of enttype_subj,
        minimum width=2cm, 
        minimum height=1.2cm, 
        text width=3cm, 
        text opacity=1, 
        align=center,
    ] (def_enttype_subj) {\LARGE{definition}};

    \node[
        shape=rectangle,
        draw=black,
        thick,
        fill=gray!50, 
        opacity=0.3, 
        below=0.8cm of evttype,
        minimum width=2cm, 
        minimum height=1.2cm, 
        text width=3cm, 
        text opacity=1, 
        align=center,
    ] (def_evttype) {\LARGE{definition}};

    \node[
        shape=rectangle,
        draw=black,
        thick,
        fill=gray!50, 
        opacity=0.3, 
        below=0.8cm of enttype_obj,
        minimum width=2cm, 
        minimum height=1.2cm, 
        text width=3cm, 
        text opacity=1, 
        align=center,
    ] (def_enttype_obj) {\LARGE{definition}};

    \draw [-, color=black, line width=1.5pt] (subject) -- node[left] {} (verb);
    \draw [-, color=black, line width=1.5pt] (verb) -- node[left] {} (object);
    \draw [-, color=black, line width=1.5pt] (enttype_subj) -- node[left] {} (evttype);
    \draw [-, color=black, line width=1.5pt] (evttype) -- node[left] {} (enttype_obj);
    \draw [-, color=black, line width=1.5pt] (def_enttype_subj) -- node[left] {} (def_evttype);
    \draw [-, color=black, line width=1.5pt] (def_evttype) -- node[left] {} (def_enttype_obj);

\end{tikzpicture}
    }
    \caption{Simple \textit{(s,v,o)}-event enhanced by fine-grained entity types for the subject and object and event type for the verb, accompanied with their definitions.}
    \label{fig:intro_approach_example}
\end{figure}

To mitigate these issues, we propose to inject fine-grained entity types and event types as external knowledge to the model. We design multiple templates to construct natural language prompts to inject types of entities (i.e. subject and object) and events (i.e. verb), together with their definitions extracted from a knowledge base. We also perform data augmentation to counteract the unbalanced label distribution. Experimental results verify our hypothesis that the model benefits from the injected knowledge. 

The main contributions of this paper are: (1) We enhance an LLM with fine-grained entity- and event knowledge for the semantic plausibility modeling problem; (2) We incorporate rich semantic knowledge of the entity type and event type from an external knowledge base into their labels; (3) We fuse these knowledge using multiple specifically designed templates; (4) We augment the dataset to deal with the unbalanced label distribution.

\section{Related Work}
\label{sec:related_work}
\subsection{Semantic Plausibility Modeling}
Semantic plausibility is a fundamental element for understanding event semantics due to the multi-faceted human intuition in the assessment \citep{resnik1996selectional} and the infrequency, non-typicality and non-preference of plausible and implausible events \citep{pado2007flexible}. It describes what is likely, but not necessarily attested in a given world \citep{gordon2013reporting}. In contrast to selectional preference \citep{erk2010exemplar}, which is characterized by the typicality of events, semantic plausibility is sensitive to certain properties that are not explicitly covered by selectional preference \citep{bagherinezhad2016elephants}.


Modeling semantic plausibility is the task of distinguishing events that are likely to happen and those whose occurrences are implausible. Previous efforts focus on injecting world knowledge about entity properties (\citealt{forbes-choi-2017-verb,wang2018modeling}) and capturing abstractness in simple events \citep{eichel2023dataset}, which requires domain expertise for the annotation. Furthermore, entity properties such as \textit{size} and \textit{weight} may not be sufficient for the model to learn the relationship between the subject and object involved in an event. As we will show, knowledge about “what kind of action is occurring in the event” and “what kind of entities are participating in it”, which are given by the type labels and their detailed definitions, can be an effective alternative to the entity properties.

\subsection{Ultra Fine-grained Entity Typing}
Ultra Fine-grained Entity Typing (UFET) is a multi-label classification problem of predicting fine-grained semantic types for entity mentions in text \citep{choi2018ultra}. The most significant challenge attributes to the massive label space of entity types (typically over 10k classes). Existing approaches can be categorized into two lines: modeling label dependencies and type hierarchies \citep{onoe2021modeling,zuo2022type}, as well as data augmentation with distant supervision \citep{dai2021ultra,zhang2022denoising,li2022ultra}. \citet{feng2023calibrated} 
 propose CASENT, a sequence-to-sequence system that predicts ultra-fine entity types 
using probability calibration. The model is trained to predict a ground-truth entity type given an input entity mention in an auto-regressive manner. During inference, a calibration module computes a calibrated confidence score for each predicted candidate type label of the entity mention. The system then selects the final predictions from the candidate labels using a threshold on the confidence scores. CASENT achieves new state-of-the-art performance on the UFET dataset \citep{choi2018ultra}.


\subsection{Event Detection}
An event is defined as an occurrence of an action that causes the change of a state \citep{li2022survey}.
Events are represented in various ways in different studies.
In early years, an event is defined either as a proposition of subject and predicate in studies on temporal news comprehension \citep{filatova2001assigning}, or as a \textit{(predicate, dependency)}-pair in studies on script learning \citep{chambers2008unsupervised}. Later, \citet{balasubramanian2013generating} represent events by \textit{(subject, relation, object)}-triples for event schema induction.
Recent studies in information extraction define an event as a more complex structure form consisting of event trigger, event type, event arguments and argument roles \citep{li2022survey}. The trigger is the core unit of an event, typically the verb serving as the predicate. The event type describes the representative feature of the event and is usually the type of the trigger. Event arguments are participants involving in the event and other details about the event, such as time and place.

Event detection is the task of identifying event triggers in a given event mention and classifying them into event types. Although multiple benchmarks are proposed for event detection \citep{grishman2005nyu,wang2020maven}, they are limited to the ranges of topics and suffer from small data sizes. \citet{li2023glen} release GLEN, a new event detection dataset with a larger data size and a wider coverage of type labels.



\section{Task Definition}
\label{sec:task}
We formulate the semantic plausibility modeling as a binary sequence classification problem. Given a knowledge-enhanced event mention $x$ as prompt, the model should predict whether it is \textit{plausible (1)} or \textit{implausible (0)}, i.e.
\[
\hat{y} = \argmax_{y \in \{0, 1\}} P(y | h(x))
\]
where $h(x)$ is the output of model's last hidden layer for $x$.



\section{Methods}
\label{sec:methods}
\begin{table*}[htbp]
    \centering
    \begin{tabular}{l|ccc}
        \toprule
        \textbf{subj / obj} 
            & \multicolumn{3}{c}{\textbf{entity types}} \\ 
            \cline{2-4}
            & \textbf{wd\_qid} & \textbf{wd\_label} & \textbf{description} \\
        \midrule
        Trader & \begin{tabular}{@{}c@{}} Q215627 \\ Q43845 \\ Q1424605 \\ Q702269 \\ Q131524 \end{tabular} & \begin{tabular}{@{}c@{}} person \\ businessperson \\ trader \\ professional \\ entrepreneur \end{tabular} & \begin{tabular}{@{}c@{}} being that has certain capacities ... \\ person involved in activities for ... \\ businessperson who exchanges ... \\ person who is paid to undertake ... \\ individual who organizes and ... \end{tabular} \\
        \midrule
        strategy & \begin{tabular}{@{}c@{}} Q131841 \\ Q151885 \\ Q1371819 \end{tabular} & \begin{tabular}{@{}c@{}} idea \\ concept \\ plan \end{tabular} & \begin{tabular}{@{}c@{}} mental image or concept \\ semantic unit understood in ... \\ outline of a strategy for ... \end{tabular} \\
        \bottomrule
    \end{tabular}
    \caption{An example UFET prediction for the subject and object in the event \textit{(trader, ensures, strategy)}.}
    \label{tab:ufet_example}
\end{table*}

\begin{table*}[htbp]
    \centering
    \begin{tabular}{l@{~}|ccc}
        \toprule
        \textbf{event trigger} 
            & \multicolumn{3}{c}{\textbf{event type}} \\
            \cline{2-4}
            & \textbf{xpo\_node} & \textbf{name} & \textbf{description} \\
        \midrule
        robs & DWD\_Q53706 & robbery & taking or attempting to take ... \\
        accusation & DWD\_Q19357312 & accusation & act of accusing or charging ... \\
        \bottomrule
    \end{tabular}
    \caption{An example event detection prediction for the event \textit{(option, robs, accusation)}.}
    \label{tab:dt_example}
\end{table*}

\subsection{Data Augmentation}



To address the issue of unbalanced label distribution towards plausible events in one of our datasets, we employ a data augmentation strategy to increase the number of implausible events. This involves enriching the binary-class variant of the dataset with unduplicated implausible events randomly sampled from the multi-class variant of the dataset which also provides additional binary plausibility labels. 



\subsection{Ultra Fine-grained Entity Typing}

We perform UFET to obtain fine-grained entity types and their definitions using \textbf{CASENT} \citep{feng2023calibrated}. For each event triple $(s,v,o)$, we produce two sentences, one with the text span for the subject \textsl{s} indicated using the special tokens <M> and </M>, the other with the text span for the object \textsl{o} indicated using the same special tokens. Then, CASENT predicts a set of fine-grained entity types for \textsl{s} in the first sentence and for \textsl{o} in the second sentence. After that, we extract the definitions (“description”) for the predicted type labels from a knowledge base (KB) built on WikiData\footnote{\url{https://wikidata.org}}.


For example, given the event \textit{(trader, ensures, strategy)}, we produce two sentences “<M> Trader </M> ensures strategy.” and “Trader ensures <M> strategy </M>.” as the inputs to CASENT. The system outputs 5 entity types for the subject \textit{trader} and 3 entity types for the object \textit{strategy} together with their definitions as shown in Table \ref{tab:ufet_example}.

\subsection{Event Detection}
For each natural language sentence built from an event triple $(s,v,o)$, we identify the event trigger (which is usually $v$) and predict its type label $t_{v}$ using the event detection model \textbf{CEDAR} accompanied with GLEN \citep{li2023glen}. Furthermore, we extract the definition of the event type label from the KB accompanied with the dataset which is also built on WikiData. An event can have multiple event triggers. Each event trigger, however, can only have one event type. 

Table \ref{tab:dt_example} shows the predictions of CEDAR for the event \textit{(option, robs, accusation)} as an example. The model identifies two event triggers \textit{robs} and \textit{accusation}, assigned respectively with the type \textit{robberty} with its definition \textit{taking or attempting to take something of value by force or threat of force or by putting the victim in fear}, and the type \textit{accusation} with its definition \textit{act of accusing or charging another with a crime}. Based on the structure of simple events, we assume that the event trigger is always the verb $v$. Thus, we consider only the event type for the verb in the model's predictions.

\begin{figure*}[h]
    \centering
    \scalebox{0.7}{
        \begin{tikzpicture}[scale=1]
    \node[
        shape=rectangle, 
        rounded corners=5pt,
        fill=green!100, 
        opacity=0.3, 
        minimum width=1cm, 
        minimum height=2cm, 
        text width=1cm, 
        text opacity=1, 
        align=center,
    ] 
        (input) 
        {\Large{S}\\ \Large{V}\\ \Large{O}};

    \node[
        shape=rectangle,
        rounded corners=5pt,
        draw=black,
        thick,
        fill=yellow!60, 
        opacity=0.3, 
        minimum width=1cm, 
        minimum height=2cm, 
        above=1cm of input,
        text width=1cm, 
        text opacity=1, 
        align=center,
    ] (ufet) {UFET};

    \node[
        shape=rectangle,
        rounded corners=5pt,
        draw=black,
        thick,
        fill=yellow!60, 
        opacity=0.3, 
        minimum width=1cm, 
        minimum height=2cm, 
        below=1cm of input,
        text width=1cm, 
        text opacity=1, 
        align=center,
    ] (ed) {ED};

    \node[
        shape=rectangle, 
        rounded corners=5pt, 
        draw=black,
        thick,
        fill=brown!30, 
        minimum width=1cm, 
        minimum height=2cm, 
        right=1.3cm of input,
        text width=1cm, 
        align=center,
        ] (template) {\rotatebox{-90}{\textsc{\Large{Template}}}};

    \node[
        shape=rectangle, 
        rounded corners=5pt, 
        fill=green!100, 
        opacity=0.3, 
        right=0.7cm of template,
        minimum width=1cm, 
        minimum height=2cm, 
        text width=1cm, 
        text opacity=1, 
        align=center,
        ] (prompt) {\rotatebox{-90}{\Large{Prompt}}};

    \node[
        right=0.7cm of prompt,
    ] (model) {\input{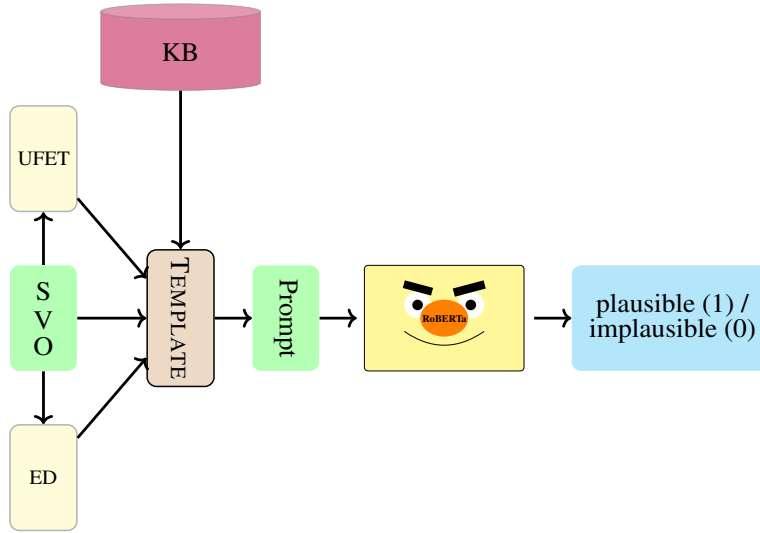}};

    \node[
        shape=rectangle, 
        rounded corners=5pt, 
        fill=cyan!100, 
        opacity=0.3, 
        right=0.7cm of model,
        minimum width=3.5cm, 
        minimum height=2cm,
        text width=3.5cm, 
        text opacity=1, 
        align=center,
        ] (output) {\Large{plausible (1) /\\ implausible (0)}};

    \node[
        shape=cylinder, 
        draw,
        fill=purple!100, 
        shape aspect=0.3,
        opacity=0.3, 
        above=3cm of template.north,
        anchor=west,
        minimum width=3cm, 
        minimum height=1cm,
        text width=1cm, 
        text opacity=1, 
        align=center,
        cylinder uses custom fill, 
        cylinder body fill=purple!100, 
        cylinder end fill=purple!50,
        rotate=90,
    ] (knowledge) {\rotatebox{-90}{\Large{KB}}};

    \draw [->, color=black, line width=1.5pt] (input) -- node[left] {} (template);
    \draw [->, color=black, line width=1.5pt] (input) -- node[left] {} (ufet);
    \draw [->, color=black, line width=1.5pt] (input) -- node[left] {} (ed);
    \draw [->, color=black, line width=1.5pt] (ufet) -- node[left] {} (template);
    \draw [->, color=black, line width=1.5pt] (ed) -- node[left] {} (template);
    \draw [->,color=black,line width=1.5pt] (template) -- node[left] {} (prompt);
    \draw [->,color=black,line width=1.5pt] (prompt) -- node[left] {} (model);
    \draw [->,color=black,line width=1.5pt] (model) -- node[left] {} (output);
    \draw[->, color=black,line width=1.5pt] (knowledge) -- node[left] {} (template);

\end{tikzpicture}
    }
    \caption{System architecture.}
    \label{fig:system_architecture}
\end{figure*}

\subsection{Template-based Prompt Engineering}
To better integrate the knowledge elements into the input, we design the following 9 templates to construct natural language prompts containing the event sentence, entity types and the event type, along with their definitions.

\begin{enumerate}
    \item TEMPLATE\_SENT = [EVT]\{sent\}[/EVT]
    \item TEMPLATE\_SUBJ\_BASIC = The subject “\{subj\}” has type [STYPE]\{stype\}[/STYPE],\\ which means [DEF]\{stype\_desc\}[/DEF].
    \item TEMPLATE\_SUBJ\_EXTEND = It can also have type [STYPE]\{stype\}[/STYPE], which means [DEF]{stype\_desc}[/DEF].
    \item TEMPLATE\_SUBJ\_UNK = The subject “\{subj\}” has an unknown type.
    \item TEMPLATE\_VERB = The verb “\{verb\}” has type [ETYPE]{etype}[/ETYPE], which means [DEF]{etype\_desc}[/DEF].
    \item TEMPLATE\_VERB\_UNK = The verb “\{verb\}” has an unknown type.
    \item TEMPLATE\_OBJ\_BASIC = The object “\{obj\}” has type [OTYPE]{otype}[/OTYPE], which means [DEF]{otype\_desc}[/DEF].
    \item TEMPLATE\_OBJ\_EXTEND = It can also have type [OTYPE]{otype}[/OTYPE], which means [DEF]{otype\_desc}[/DEF].
    \item TEMPLATE\_OBJ\_UNK = The object “\{obj\}” has an unknown type.
\end{enumerate}

\section{Experiments}
\label{sec:experiments}
\subsection{Data}

Our experiments are conducted on \textbf{PEP-3K} \citep{wang2018modeling} and \textbf{PAP} \citep{eichel2023dataset}. We use the values in the column \textit{label} as ground truth labels. 
The train- and dev sets of the both datasets are merged for the fine-tuning, each resulting in 4911 and 614 examples. The model is evaluated separately on the test sets of the two datasets, each containing 307 and 308 examples.

\subsection{Preliminary Study}

In our initial exploration, we investigate the statistic and linguistic characteristics of two datasets to glean insights into their composition and semantic nuances. Our analysis encompasses an examination of the distribution of examples across training, development, and test splits for each dataset. Additionally, we employ word clouds as a visual tool to highlight the most frequently occurring words tied to both plausible and implausible events, offering a vivid portrayal of the datasets' lexical landscapes. The outcomes of this investigation are presented in Appendix \ref{sec:appendix}.

\begin{table*}[htbp]
    \centering
    \begin{tabular}{l@{~}|ccccc|ccccc}
        \toprule
        \multirow{2.5}{*}{\textbf{Model}}
            & \multicolumn{5}{c}{\textbf{PAP}}
            & \multicolumn{5}{c}{\textbf{PEP-3K}} \\
            \cmidrule(lr){2-6}\cmidrule(lr){7-11}
            & AUC & P & R & F1 & Acc
            & AUC & P & R & F1 & Acc \\
        \midrule
        $\textsc{RoBERTa}_{evt+ent,ft}$ & 0.659 & 0.717 & 0.526 & 0.607 & 0.659 & \textbf{0.883} & \textbf{0.850} & \textbf{0.928} & \textbf{0.888} & \textbf{0.883} \\
        $\textsc{RoBERTa}_{evt,ft}$ & 0.636 & 0.640 & \textbf{0.623} & \textbf{0.632} & 0.636 & 0.844 & 0.793 & \textbf{0.928} & 0.855 & 0.844 \\
        $\textsc{RoBERTa}_{ent,ft}$ & \textbf{0.666} & \textbf{0.763} & 0.481 & 0.590 & \textbf{0.666} & 0.840 & 0.838 & 0.843 & 0.840 & 0.840 \\
        \midrule
        $\textsc{RoBERTa}_{bs,0-shot}$ & 0.532 & 0.564 & 0.286 & 0.379 & 0.532 & 0.500 & 0.500 & 0.137 & 0.215 & 0.502 \\
        $\textsc{RoBERTa}_{bs,ft}$ & 0.646 & 0.737 & 0.455 & 0.562 & 0.646 & 0.791 & 0.830 & 0.732 & 0.778 & 0.792 \\
        \bottomrule
    \end{tabular}
    \caption{Semantic plausibility modeling results. For PAP, injecting entity type leads to the best AUC. For PEP-3K, injecting both event type and entity type significantly improves all metrics. For both datasets, injecting event type improves the recall but reduces the precision, while injecting entity type improves the precision but decreases the recall. \textit{evt+ent}: event type and entity type, \textit{evt}: event type, \textit{ent}: entity type, \textit{bs}: baseline, \textit{ft}: fine-tune. 
    }
    \label{tab:eval_results}
\end{table*}

We utilize the python library Gensim\footnote{\url{https://github.com/piskvorky/gensim}} library to calculate semantic similarities between the sets of most frequently appearing words in the context of plausible and implausible events. This comparative analysis yields intriguing findings: the datasets exhibit contrasting semantic profiles, with the PAP dataset showcasing a pronounced dissimilarity between plausible and implausible terms, while the PEP-3K dataset reveals a striking similarity among its terms. This distinction not only sheds light on the inherent linguistic patterns within the datasets but also correlates with our subsequent experimental observations, where classifiers tend to exhibit enhanced performance on the PEP-3K dataset as opposed to the PAP dataset. Figure \ref{fig:word_clouds} depicts words similarities between top plausible and implausible words in the datasets.

\subsection{Model Implementation}
We fine-tune \textbf{RoBERTa-large}\footnote{\url{https://huggingface.co/roberta-large}} \citep{liu2019roberta} 
on the merged dataset enhanced by entity type and event type knowledge (\textsc{Evt+Ent}). As ablation study, we also fine-tune the model on the data with only event type knowledge injected (\textsc{Evt}), as well as with only entity type knowledge injected (\textsc{Ent}). Figure \ref{fig:system_architecture} illustrates the architecture of our system.

\subsubsection{Baselines}
We compare our approach with two baselines, including (1) a zero-shot inference baseline (\textsc{BS-0-shot}); and (2) RoBERTa fine-tuned on the event mentions without knowledge injection (\textsc{BS-ft}).

\subsection{Hyperparameters}
For \textsc{Evt+Ent} and \textsc{Ent}, we fine-tune the model for 10 epochs with batch size 16, using AdamW optimizer \citep{loshchilov2017decoupled}. The training procedure has an initial learning rate of 1e-5, a weight decay of 0.01 and warm-up steps 10. The \textsc{Evt} group has almost the same set of hyperparameter values, except the warm-up steps being set to 100.

The baseline \textsc{BS-ft} is fine-tuned for 10 epochs with batch size 8 using the AdamW optimizer. The learning rate is 1e-5, with a weight decay of 0.01 and warm-up steps 10.

\subsection{Evaluation Metrics}
While previous works prefer to use accuracy (Acc) for the evaluation, we also report precision (P), recall (R), F1 and Area Under the Curve (AUC).

\section{Results}
\label{sec:results}

The experimental results are shown in Table \ref{tab:eval_results}. $\textsc{RoBERTa}_{evt+ent}$ achieves the highest performance on PEP-3K with respect to all metrics, indicating the effectiveness of event type and entity type knowledge for improving semantic plausibility understanding. On PAP, $\textsc{RoBERTa}_{evt+ent}$ surprisingly achieves a lower AUC score than $\textsc{RoBERTa}_{ent}$ which injects only the entity type knowledge. This shows the limitation of our approach on understanding semantic plausibility of events of different abstractness degrees. For both datasets, injecting event type knowledge improves the recall but reduces the precision, while injecting entity type knowledge increases the precision but suppresses the recall. The results of the baselines indicate that fine-tuning is generally better than zero-shot inference.

\section{Qualitative Analysis}
\label{sec:analysis}
We observe 105 wrong predictions from the 308 examples in the PAP test set. 68 examples are assigned with an \textit{unknown} type by the model. An instance is present in Table \ref{tab:unknown_evt_type}. Furthermore, 50 of the wrong predictions are assigned with a trivial entity type \textit{entity}. As shown in Table \ref{tab:trivial_ent_type}, the co-occurrence of such trivial entity types and the unknown event type may be harmful for the model to get a precise understanding of the event plausibility.

\begin{table*}[htbp]
    \centering
    \begin{tabular}{l|c}
        \toprule
        \textsc{Event} & (trader, ensures, strategy) \\
        \midrule
        \textsc{Prompt} & \begin{tabular}{@{}l@{}} [EVT] Trader ensures strategy. [/EVT]\\The subject “Trader” has type [STYPE]person[/STYPE], which means [DEF]being that\\ has certain capacities or attributes constituting personhood (avoid use with P31; use Q5\\ for humans)[/DEF]. It can also have type [STYPE]businessperson[/STYPE], which me-\\ans [DEF]person involved in activities for the purpose of generating revenue[/DEF]. It\\ can also have type [STYPE]trader[/STYPE], which means [DEF]businessperson who\\ exchanges stocks, bonds and other such financial instruments[/DEF]. It can also have\\ type [STYPE]professional[/STYPE], which means [DEF]person who is paid to under-\\take a specialized set of tasks and to complete them for a fee[/DEF]. It can also have\\ type [STYPE]entrepreneur[/STYPE], which means [DEF]individual who organizes and\\ operates a business[/DEF].\\ The verb “ensures” has an \textcolor{red}{unknown type}.\\ The object “strategy” has type [OTYPE]idea[/OTYPE], which means [DEF]mental\\ image or concept[/DEF]. It can also have type [OTYPE]concept[/OTYPE], which\\ means [DEF]semantic unit understood in different ways, e.g. as mental representation,\\ ability or abstract object[/DEF]. It can also have type [OTYPE]plan[/OTYPE], which\\ means [DEF]outline of a strategy for achievement of an objective[/DEF]. \end{tabular} \\
        \midrule
        \begin{tabular}{@{}l@{}} \textsc{Predicted}\\ \textsc{Label} \end{tabular} & 0 \\
        \midrule
        \begin{tabular}{@{}l@{}} \textsc{True}\\ \textsc{Label} \end{tabular} & 1 \\
        \bottomrule
    \end{tabular}
    \caption{A wrong prediction in PAP made by $\textsc{RoBERTa}_{evt+ent}$. The event type is predicted as \textit{unknown}.}
    \label{tab:unknown_evt_type}
\end{table*}

\begin{table*}[htbp]
    \centering
    \begin{tabular}{l|c}
        \toprule
        \textsc{Event} & (hook, wins, role) \\
        \midrule
        \textsc{Prompt} & \begin{tabular}{@{}l@{}} [EVT] Hook wins role. [/EVT]\\The subject “Hook” has type [STYPE]concept[/STYPE], which means [DEF]semantic unit\\ understood in different ways, e.g. as mental representation, ability or abstract object[/DEF].\\ It can also have type [STYPE]idea[/STYPE], which means [DEF]mental image or\\ concept[/DEF]. It can also have type [STYPE]\textcolor{orange}{entity}[/STYPE], which means [DEF]anything\\ that can be considered, discussed, or observed[/DEF]. It can also have type [STYPE]\\hook[/STYPE], which means [DEF]object for hanging, fishing etc.[/DEF].\\ The verb “wins” has an \textcolor{red}{unknown type}.\\ The object “role” has type [OTYPE]\textcolor{orange}{entity}[/OTYPE], which means [DEF]anything that can\\ be considered, discussed, or observed[/DEF]. It can also have type [OTYPE]role[/OTYPE],\\ which means [DEF]set of behaviours, rights, obligations, beliefs, and norms expected from\\ an individual that has a certain social status[/DEF]. \end{tabular} \\
        \midrule
        \begin{tabular}{@{}l@{}} \textsc{Predicted}\\ \textsc{Label} \end{tabular} & 1 \\
        \midrule
        \begin{tabular}{@{}l@{}} \textsc{True}\\ \textsc{Label} \end{tabular} & 0 \\
        \bottomrule
    \end{tabular}
    \caption{A wrong prediction in PAP made by $\textsc{RoBERTa}_{evt+ent}$. The entities are assigned with a trivial type \textit{entity}.}
    \label{tab:trivial_ent_type}
\end{table*}

\section{Conclusion}
\label{sec:conclusion}
This paper proposes to enhance a large language model with information about fine-grained entity types, event types and their definitions extracted from an external knowledge base. We design templates to integrate these knowledge into the model's input and mitigate the unbalanced label distribution via data augmentation. In addition, we adapt the task to real world scenarios by converting simple events to natural langauge sentences. The experimental results on PEP-3K shows that the model's performance on the task benefits from injecting these knowledge. However, there is still big room of improvement on the PAP dataset. This unfolds the need to address abstractness in modeling semantic plausibility.

\section*{Limitations}
\label{sec:limitations}
Our experiments are conducted on simple events in the form of \textit{(s,v,o)}-triples. However, events in the real world scenario usually comprise much more information and are represented in much more complex ways. Furthermore, we do not investigate argument roles in this paper since CEDAR focuses on event detection rather than event extraction, which involves also the identification of event arguments and prediction of their roles. Last but not least, the datasets only contain events expressed in texts, while events can also be depicted by other modalities, such as images and videos. In the future research, we will work on more complex event structure, build an event extraction system that identifies event arguments and classifies argument roles in addition to detecting events from other modalities.




\bibliography{anthology,custom}
\bibliographystyle{msp_natbib}

\appendix
\label{sec:appendix}
\section{Appendix}
\subsection{Most Frequent Words}
\begin{figure*}[htbp]
    \centering
        \subfigure[PAP (train), plausible]
        {
            \begin{minipage}[t]{0.4\linewidth}
            \centering
            \includegraphics[width=1\textwidth]{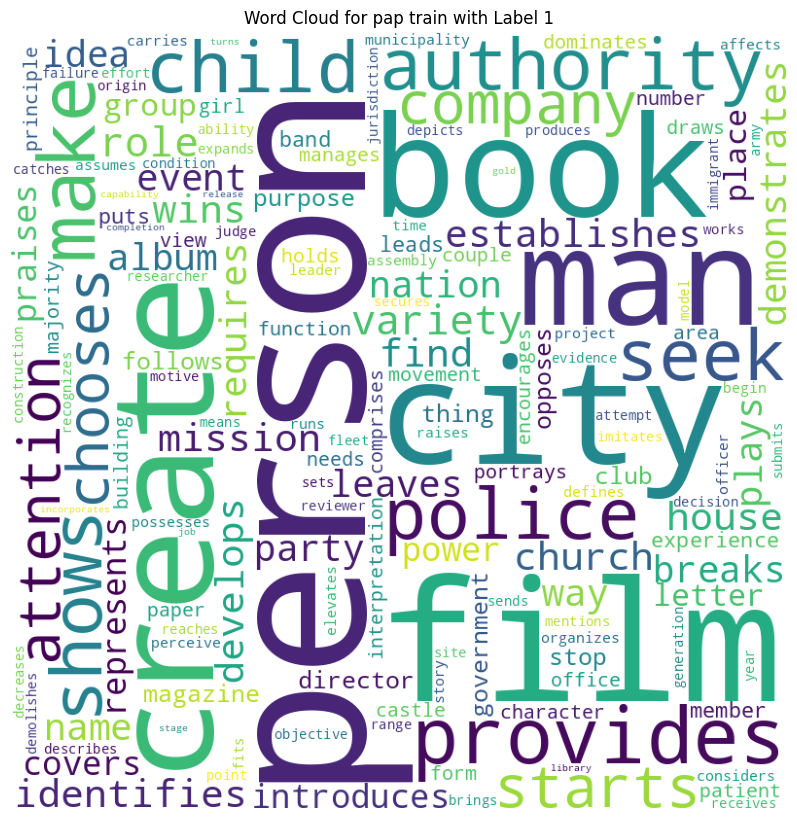}
            \end{minipage}
        }\hspace{3mm}
        \subfigure[PAP (train), implausible]
        {
            \begin{minipage}[t]{0.4\linewidth}
            \centering
            \includegraphics[width=1\textwidth]{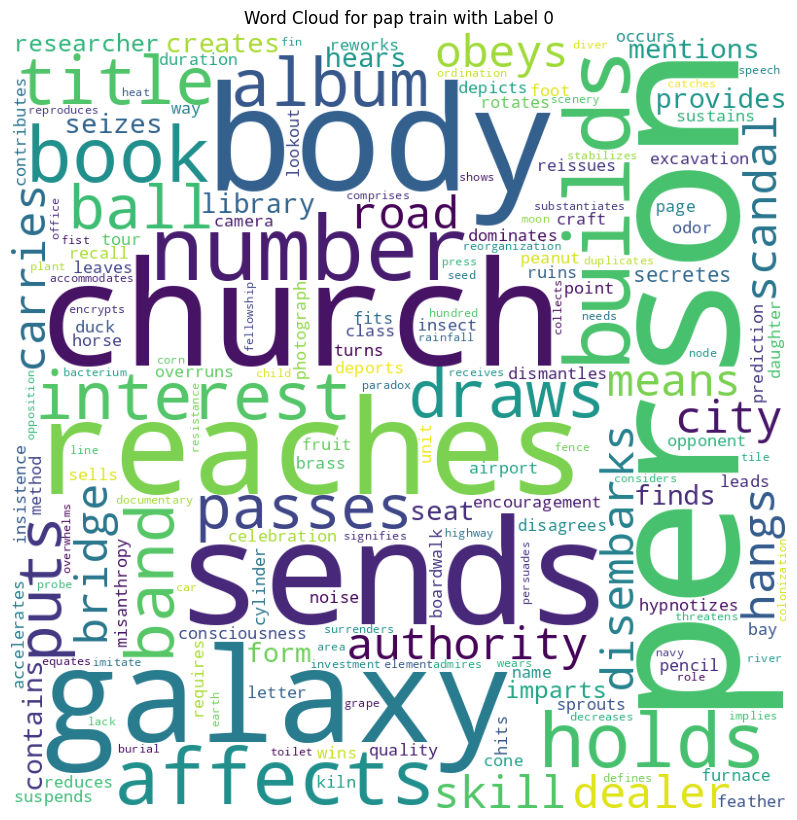}
            \end{minipage}
        }\hspace{3mm}

        \subfigure[PEP-3K (train), plausible]
        {
            \begin{minipage}[t]{0.4\linewidth}
            \centering
            \includegraphics[width=1\textwidth]{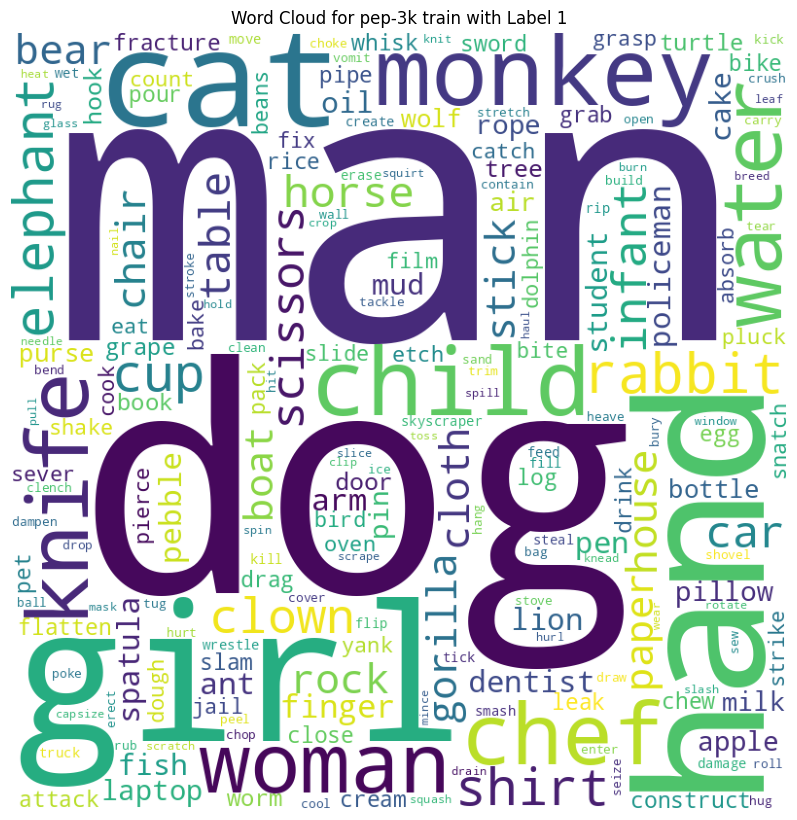}
            \end{minipage}
        }\hspace{3mm}
        \subfigure[in PEP-3K (train), implausible]
        {
            \begin{minipage}[t]{0.4\linewidth}
            \centering
            \includegraphics[width=1\textwidth]{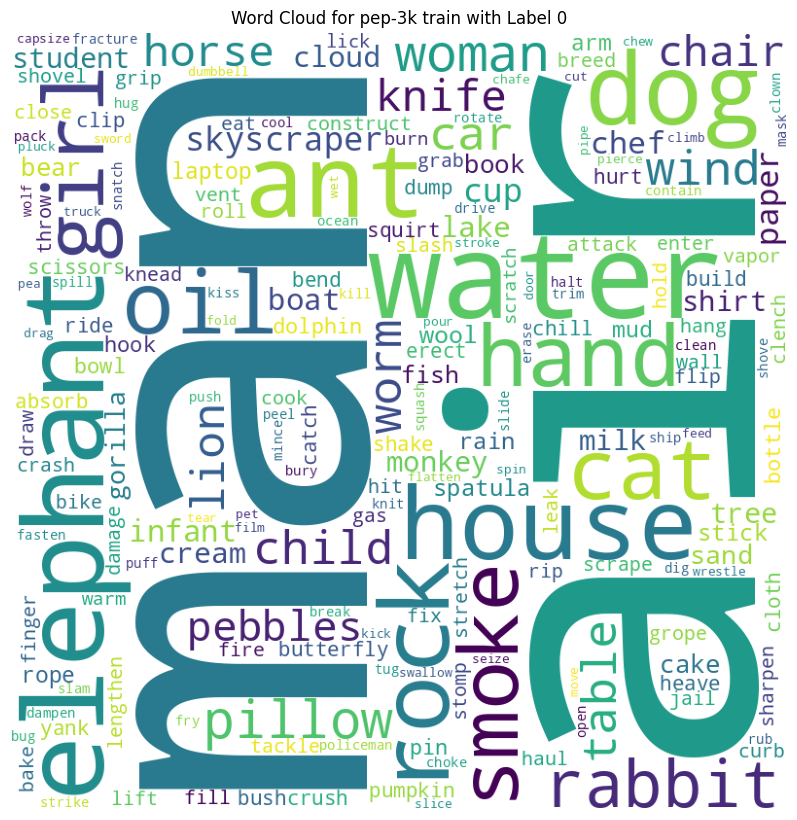}
            \end{minipage}
        }\hspace{3mm}
    \caption{Word clouds of the most frequent words associated with the labels in PEP-3K train split.}
    \label{fig:word_clouds}
\end{figure*}

Figure \ref{fig:word_clouds} illustrates the word clouds of the most frequent words associated with labels in different dataset splits.




\subsection{Semantic Similarity}
\begin{figure*}[h]
    \centering
    \includegraphics[width=0.8\textwidth]{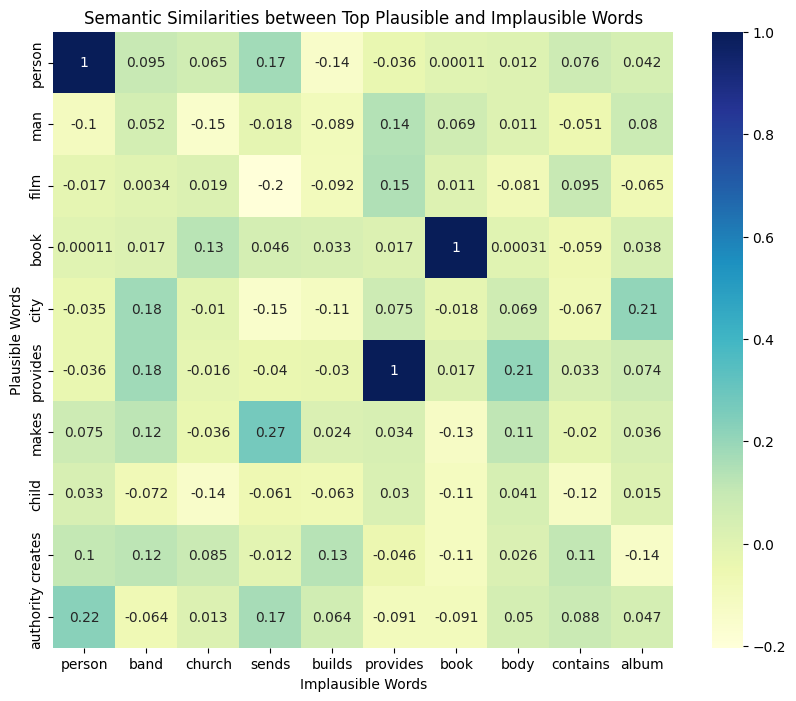}
    \caption{Word similarity between top plausible words and implausible words in PAP}
    \label{fig:word_sim_pap}
\end{figure*}

\begin{figure*}[h]
    \centering
    \includegraphics[width=0.8\textwidth]{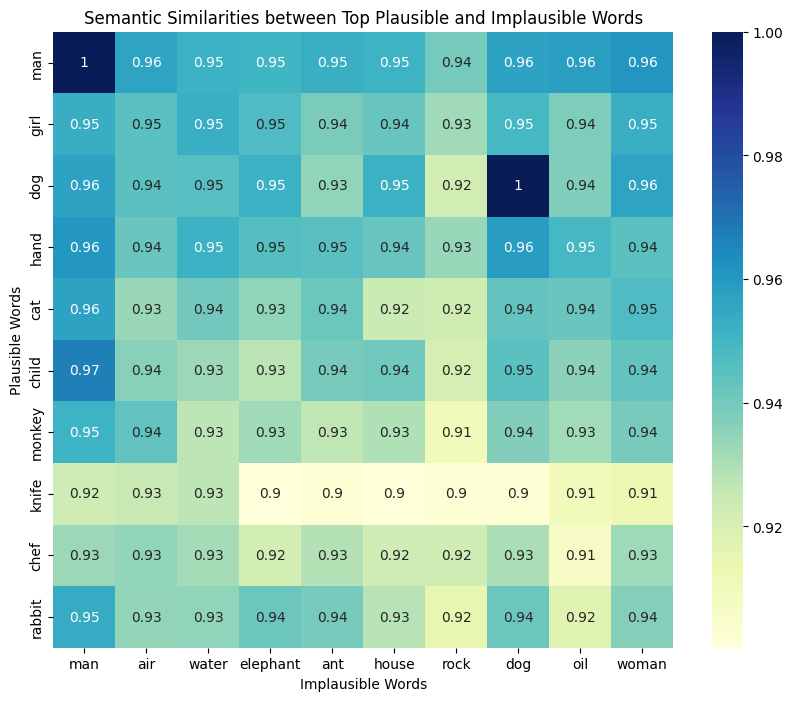}
    \caption{Word similarity between top plausible words and implausible words in PEP-3K}
    \label{fig:word_sim_pep}
\end{figure*}

Figure \ref{fig:word_sim_pap} and \ref{fig:word_sim_pep} depict words similarities between top plausible- and implausible words in the datasets.

 \end{document}